\documentclass[10pt,twocolumn,letterpaper]{article}

\usepackage[final]{cvpr}      

\usepackage{times}
\usepackage{epsfig}
\usepackage{graphicx}
\usepackage{amsmath}
\usepackage{amssymb}
\usepackage{multirow}
\usepackage{xr}
\usepackage{makecell}
\usepackage{xcolor}
\usepackage{enumerate}
\usepackage{enumitem}
\usepackage{booktabs}
\makeatletter
\usepackage[pagebackref,breaklinks,colorlinks]{hyperref}
\usepackage[capitalize]{cleveref}
\newcommand\best[1]{\textcolor{red}{\textbf{#1}}}

\newcommand\second[1]{\textcolor{cyan}{\textbf{#1}}}

\def\name {SimpleTrack}
\def\cvprPaperID{4373} 
\def\confName{CVPR}
\def\confYear{2022}
\begin{document}

\title{SimpleTrack: Understanding and Rethinking 3D Multi-object Tracking}
	
\author{Ziqi Pang$^{1}$\thanks{This work is complete during the first author's internship at TuSimple.}\qquad Zhichao Li$^{2}$\qquad Naiyan Wang$^{2}$ \\ 
	UIUC$^{1}$\qquad TuSimple$^{2}$ \\
	\tt\small ziqip2@illinois.edu, \{leeisabug, winsty\}@gmail.com}
\maketitle
	
\begin{abstract}

3D multi-object tracking (MOT) has witnessed numerous novel benchmarks and approaches in recent years, especially those under the “tracking-by-detection” paradigm. Despite their progress and usefulness, an in-depth analysis of their strengths and weaknesses is not yet available. In this paper, we summarize current 3D MOT methods into a unified framework by decomposing them into four constituent parts: pre-processing of detection, association, motion model, and life cycle management. We then ascribe the failure cases of existing algorithms to each component and investigate them in detail. Based on the analyses, we propose corresponding improvements which lead to a strong yet simple baseline: \name. Comprehensive experimental results on Waymo Open Dataset and nuScenes demonstrate that our final method could achieve new state-of-the-art results with minor modifications.
		
Furthermore, we take additional steps and rethink whether current benchmarks authentically reflect the ability of algorithms for real-world challenges. We delve into the details of existing benchmarks and find some intriguing facts. Finally, we analyze the distribution and causes of remaining failures in \name\ and propose future directions for 3D MOT. Our code is available at \url{https://github.com/TuSimple/SimpleTrack}.
		
\end{abstract}
	
\section{Introduction}

Multi-object tracking (MOT) is a composite task in computer vision, combining both the aspects of localization and identification. Given its complex nature, MOT systems generally involve numerous interconnected parts, such as the selection of detections, the data association, the modeling of object motions, etc. Each of these modules has its special treatment and can significantly affect the system performance as a whole. Therefore, we would like to ask \emph{which components in 3D MOT play the most important roles, and how can we improve them?}


Bearing such objectives, we revisit the current 3D MOT algorithms~\cite{ab3dmot, mdis, center_point, graphmot, poshmann2020factor, score_refinement_3dmot}. These methods mostly adopt the ``tracking by detection'' paradigm, where they directly take the bounding boxes from 3D detectors and build up tracklets across frames. We first break them down into four individual modules and examine each of them: pre-processing of input detections, motion model, association, and life cycle management. Based on this modular framework, we locate and ascribe the failure cases of 3D MOT to the corresponding components and discover
several overlooked issues in the previous designs.

First, we find that inaccurate input detections may contaminate the association. However, purely pruning them by a score threshold will sacrifice the recall. Second, we find that the similarity metric defined between two 3D bounding boxes need to be carefully designed. Neither distance-based nor simple IoU works well. Third, the object motion in 3D space is more predictable than that in the 2D image space. Therefore, the consensus between motion model predictions and even poor observations (low score detections) could well indicate the existence of objects.
Illuminated by these observations, we propose several simple yet non-trivial solutions. The evaluation on Waymo Open Dataset~\cite{waymo_open} and nuScenes~\cite{nuScenes} suggests that our final method ``\name'' is competitive among the 3D MOT algorithms (in Tab.~\ref{tab::wod_test_all} and Tab.~\ref{tab::nusc_test_all}).

Besides analyzing 3D MOT algorithms, we also reflect on current benchmarks. We emphasize the need for high-frequency detections and the proper handling of output tracklets in evaluation. To better understand the upper bound of our method, we further break down the remaining errors based on ID switch and MOTA metrics. We believe these observations could inspire the better design of algorithms and benchmarks.

\begin{figure*}
    \centering
    \includegraphics[width=1.9\columnwidth]{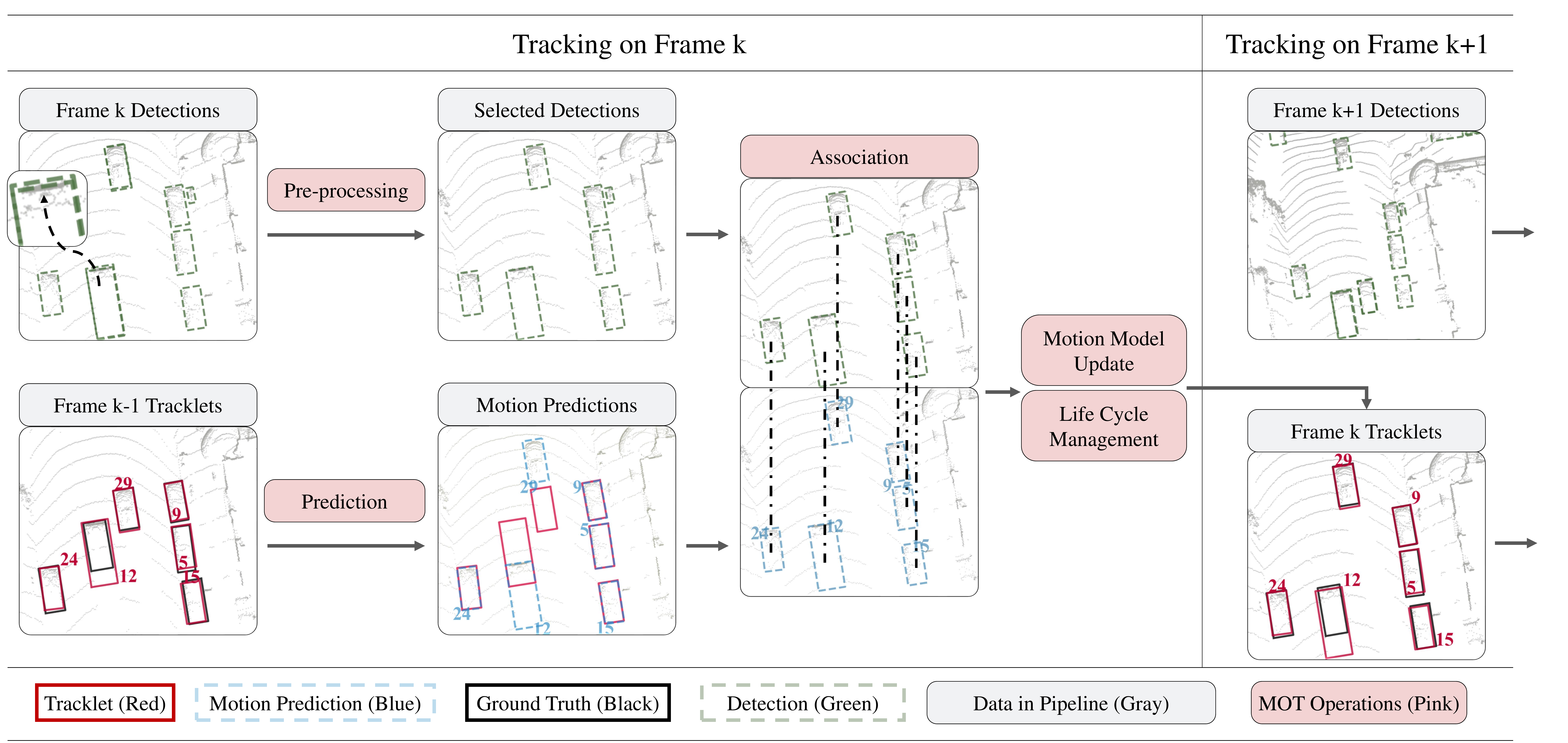}
    \caption{3D MOT pipeline. For simplicity, we only visualize the steps between frame k and frame k+1. Best view in color.}
    \vspace{-4mm}
    \label{fig::pipeline}
\end{figure*}

In brief, our contributions are as follow:
\begin{itemize}[itemsep=0mm]
    \item We decompose the pipeline of ``tracking-by-detection'' 3D MOT framework and analyze the connections between each component and failure cases.
    \item We propose corresponding treatments for each module and combine them into a simple baseline. The results are competitive on the Waymo Open Dataset and nuScenes.
    \item We also analyze existing 3D MOT benchmarks and explain the potential influences of their designs. We hope that our analyses could shed light for future research.
\end{itemize}

\section{Related Work}
Most 3D MOT methods~\cite{ab3dmot, mdis, center_point, poshmann2020factor, graphmot, score_refinement_3dmot} adopt the ``tracking-by-detection'' framework because of the strong power of detectors. We first summarize the representative 3D MOT work and then highlight the connections and distinctions between 3D and 2D MOT.

\subsection{3D MOT}
Many 3D MOT methods are composed of hand-crafted rule-based components. AB3DMOT~\cite{ab3dmot} is the common baseline of using IoU for association and a Kalman filter as the motion model. Its notable followers mainly improve on the association part: Chiu \etal~\cite{mdis} and CenterPoint~\cite{center_point} replace IoU with Mahalanobis and L2 distance, which performs better on nuScenes~\cite{nuScenes}. Some others notice the importance of life cycle management, where CBMOT~\cite{score_refinement_3dmot} proposes a score-based method to replace the ``count-based'' mechanism, and P{\"{o}}schmann \etal ~\cite{poshmann2020factor} treats 3D MOT as optimization problems on factor graphs. Despite the effectiveness of these improvements, a systematic study on 3D MOT methods is in great need, especially where these designs suffer and how to make further improvements. To this end, our paper seeks to meet the expectations.

Different from the methods mentioned above, many others attempt to solve 3D MOT with fewer manual designs.~\cite{gnn3dmot, eagermot, chui2021prob, baser2019fantrack} leverage rich features from RGB images for association and life cycle control, and Chiu \etal~\cite{chui2021prob} specially uses neural networks to handle the feature fusion, association metrics, and tracklet initialization. Recently, OGR3MOT~\cite{graphmot} follows Guillem \etal ~\cite{neural_solver} and solves 3D MOT with graph neural networks (GNN) in an end-to-end manner, focusing on the data association and the classification of active tracklets, especially.
\subsection{2D MOT}
2D MOT shares the common goal of data association with 3D MOT. Some notable attempts include probabilistic approaches~\cite{bar1990tracking, reid1979algorithm, rezatofighi2015joint, kim2015multiple}, dynamic programming~\cite{fleuret2007multicamera}, bipartite matching~\cite{bewley2016simple}, min-cost flow~\cite{zhang2008global, berclaz2011multiple}, convex optimization~\cite{tang2015subgraph, tang2016multi, zamir2012gmcp, pirsiavash2011globally}, and conditional random fields~\cite{yang2011learning}. With the rapid progress of deep learning, many methods~\cite{jiang2019graph, neural_solver, xu2020train, li2020graph, hornakova2020lifted, he2021learnable} learn the matching mechanisms and others~\cite{liang2020enhancing, peng2020tpm, liu2020gsm, lan2016online, pang2021quasi} learn the association metrics. 

Similar to 3D MOT, many 2D trackers~\cite{bergmann2019tracking, lu2020retinatrack, sadeghian2017tracking, zhang2021fairmot} also benefit from the enhanced detection quality and adopt the ``tracking-by-detection'' paradigm. However, the objects on RGB images have varied sizes because of scale variation; thus, they are harder for association and motion models. But 2D MOT can easily take advantage of rich RGB information and use appearance models~\cite{wojke2017simple, leal2016learning, sadeghian2017tracking, li2020graph}, which is not available in LiDAR based 3D MOT. In summary, the design of MOT methods should fit the traits of each modality.
\section{3D MOT Pipeline} 
\label{sec::baselines}

In this section, we decompose 3D MOT methods into the following four parts. An illustration is in Fig.~\ref{fig::pipeline} .

\paragraph{Pre-processing of Input Detections.} It pre-processes the bounding boxes from detectors and selects the ones to be used for tracking. Some exemplar operations include selecting the bounding boxes with scores higher than a certain threshold. (In ``Pre-processing'' Fig.~\ref{fig::pipeline}, some redundant bounding boxes are removed.)

\paragraph{Motion Model.}
It predicts and updates the states of objects. Most 3D MOT methods~\cite{ab3dmot, mdis, score_refinement_3dmot} directly use the Kalman filter, and CenterPoint~\cite{center_point} uses the velocities predicted by detectors from multi-frame data. (In ``Prediction'' and ``Motion Model Update'' Fig.~\ref{fig::pipeline}.)

\paragraph{Association.} 
It associates the detections with tracklets. The association module involves two steps: similarity computation and matching. The similarity measures the distance between a pair of detection and tracklet, while the matching step solves the correspondences based on the pre-computed similarities. AB3DMOT~\cite{ab3dmot} proposes the baseline of using IoU with Hungarian algorithm, while Chiu \etal~\cite{mdis} uses Mahalanobis distance and greedy algorithm, and CenterPoint~\cite{center_point} adopts the L2 distance. (In ``Association'' Fig.~\ref{fig::pipeline}.)

\paragraph{Life Cycle Management.}
It controls the ``birth'', ``death'' and ``output'' policies. ``Birth'' determines whether a detection bounding box will be initialized as a new tracklet; ``Death'' removes a tracklet when it is believed to have moved out of the attention area; ``Output'' decides whether a tracklet will output its state. Most of the MOT algorithm adopts a simple count-based rule~\cite{ab3dmot, mdis, center_point}, and CBMOT~\cite{score_refinement_3dmot} improves birth and death by amending the logic of tracklet confidences. (In ``Life Cycle Management'' Fig.~\ref{fig::pipeline}.)
\section{Analyzing and Improving 3D MOT}
\label{sec::ablation}

In this section, we analyze and improve each module in the 3D MOT pipeline. For better clarification, we ablate the effects of every modification by removing it from the final variant of \name. By default, the ablations are all on the validation split with the CenterPoint~\cite{center_point} detection. We also provide additive ablation analyses and the comparison with other methods in Sec.~\ref{subsec::summary_improvements}. 

\subsection{Pre-processing} 
\label{subsec::initialization_model}

\begin{figure}
	\centering
	\includegraphics[width=1.0\columnwidth]{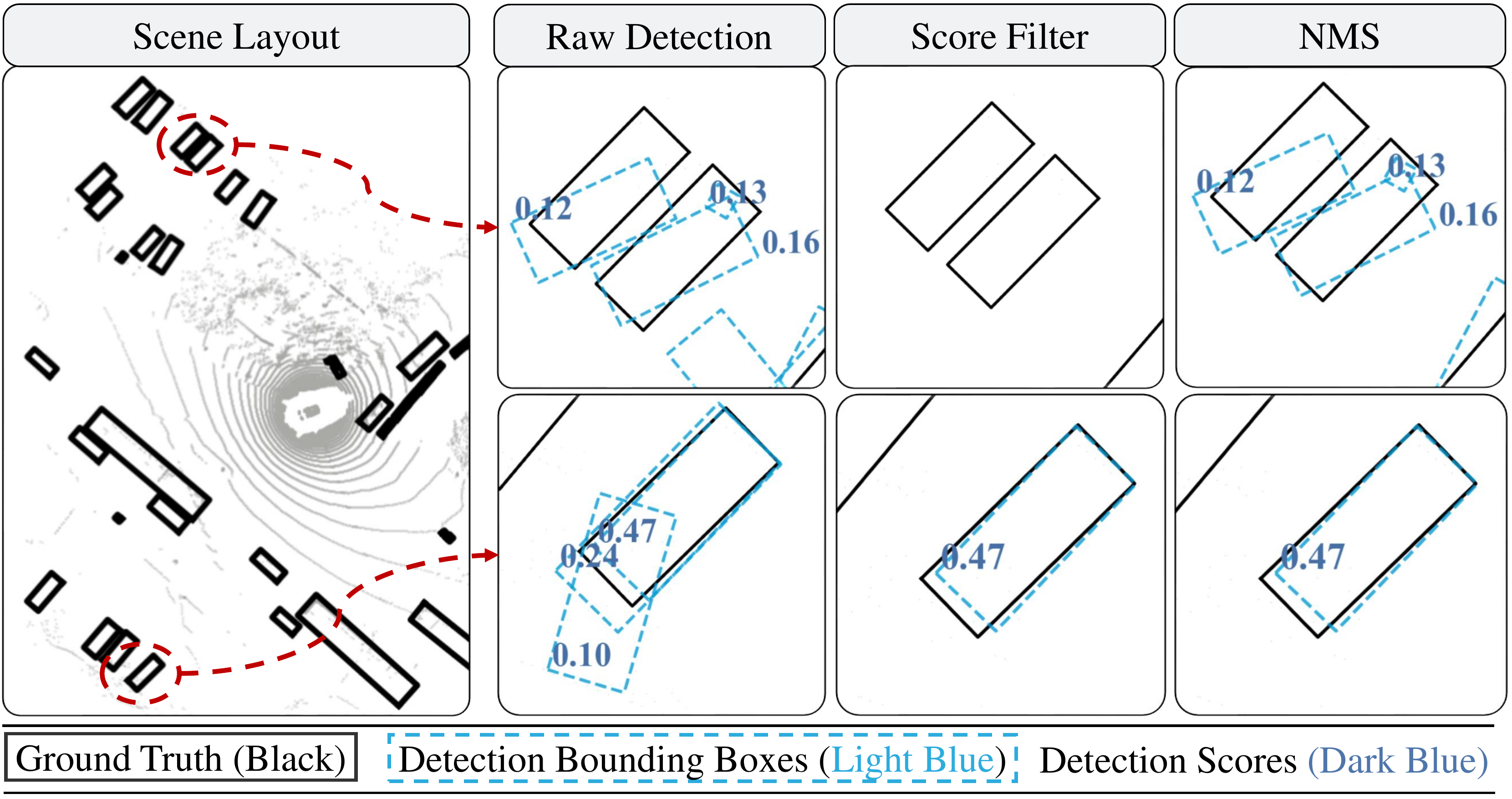}
	\caption{Comparison between score filtering and NMS. To remove the redundant bounding boxes on row 2, score filtering needs at least a 0.24 threshold, but this will eliminate the detections on row 1. However, NMS can well satisfy both by removing the overlapping on row 2 and maintaining the recall on row 1.}
	\label{fig::nms}
\end{figure}
To fulfill the recall requirements, current detectors usually output a large number of bounding boxes with scores roughly indicating their quality. However, if these boxes are treated equally in the association step of 3D MOT, the bounding boxes with low quality or severe overlapping may deviate the trackers to select the inaccurate detection for extending or forming tracklets (as in the ``raw detection'' of Fig.~\ref{fig::nms}). Such a gap between the detection and MOT task needs careful treatment.

3D MOT methods commonly use confidence scores to filter out the low-quality detections and improve the precision of input bounding boxes for MOT. However, such an approach may be detrimental to the recall as they directly abandon the objects with poor observations (top row in Fig.~\ref{fig::nms}). It is also especially harmful to metrics like AMOTA, which needs the tracker to use low score bounding boxes to fulfill the recall requirements.

\begin{table}
	\centering
	\resizebox{0.75\linewidth}{!} 
	{
		\begin{tabular}{@{\hspace{2.0mm}}c|@{\hspace{2.0mm}}c@{\hspace{2.0mm}}c@{\hspace{2.0mm}}c@{\hspace{2.0mm}}c}
			\toprule
			NMS & AMOTA$\uparrow$ & AMOTP$\downarrow$ & MOTA$\uparrow$ & IDS $\downarrow$ \\
			\midrule\midrule
			$\times$ & 0.673 & 0.574 & 0.581 & 557 \\
			\checkmark & \textbf{0.687} & \textbf{0.573} & \textbf{0.592} & \textbf{519} \\
			\bottomrule
		\end{tabular}
	}
	\vspace{-2mm}
	\caption{Ablation for NMS on nuScenes.}
	\label{tab::nuscenes_data_preprocessor}
	\vspace{-4mm}
\end{table}

\begin{table}
	\centering
	\resizebox{1.0\linewidth}{!} 
	{
		\begin{tabular}{@{\hspace{1.5mm}}c|@{\hspace{2.0mm}}c@{\hspace{2.0mm}}c@{\hspace{2.0mm}}c|@{\hspace{2.0mm}}c@{\hspace{2.0mm}}c@{\hspace{2.0mm}}c}
			\toprule
			\multirow{2}{*}{NMS} & \multicolumn{3}{c}{Vehicle} & \multicolumn{3}{c}{Pedestrian} \\
			\cmidrule{2-7} & MOTA$\uparrow$ & MOTP$\downarrow$ & IDS(\%)$\downarrow$  & MOTA$\uparrow$ & MOTP$\downarrow$ & IDS(\%)$\downarrow$  \\
			\midrule\midrule
			$\times$ & 0.5609 & \textbf{0.1681} & 0.09 & 0.4962 & \textbf{0.3090} & 5.00 \\
			\checkmark & \textbf{0.5612} & \textbf{0.1681} & \textbf{0.08} & \textbf{0.5776} & 0.3125 & \textbf{0.42} \\
			\bottomrule
		\end{tabular}
	}
	\vspace{-2mm}
	\caption{Ablation for NMS on WOD.}
	\label{tab::wod_ablate_data_preprocessor}
	\vspace{-4mm}
\end{table}

To improve the precision without significantly decreasing the recall, our solution is simple and direct: we apply stricter non-maximum suppression (NMS) to the input detections. As shown in the right of Fig.~\ref{fig::nms}, the NMS operation alone can effectively eliminate the overlapped low-quality bounding boxes while keeping the diverse low-quality observations, even on regions like sparse points or occlusion. \emph{Therefore, by adding NMS to the pre-processing module, we could roughly keep the recall, but greatly improves the precision and benefits MOT.} 

On WOD, our stricter NMS operation removes 51\% and 52\% bounding boxes for vehicles and pedestrians and nearly doubles the precision: 10.8\% to 21.1\% for vehicles, 5.1\% to 9.9\% for pedestrians. At the same time, the recall drops relatively little from 78\% to 74\% for vehicles and 83\% to 79\% for pedestrians. According to  Tab.~\ref{tab::nuscenes_data_preprocessor} and Tab.~\ref{tab::wod_ablate_data_preprocessor}, this largely benefits the performance, especially on the pedestrian (right part of Tab.~\ref{tab::wod_ablate_data_preprocessor}), where the object detection task is harder.

\subsection{Motion Model} 
\label{subsec::motion_model}

\begin{table}
	\centering
	\resizebox{1.0\linewidth}{!} 
	{
		\begin{tabular}{@{\hspace{2.0mm}}c|@{\hspace{2.0mm}}c@{\hspace{2.0mm}}c@{\hspace{2.0mm}}c|@{\hspace{2.0mm}}c@{\hspace{2.0mm}}c@{\hspace{2.0mm}}c}
			\toprule
			\multirow{2}{*}{Method} & \multicolumn{3}{c}{Vehicle} & \multicolumn{3}{c}{Pedestrian} \\
			\cmidrule{2-7} & MOTA$\uparrow$ & MOTP$\downarrow$ & IDS(\%)$\downarrow$  & MOTA$\uparrow$ & MOTP$\downarrow$ & IDS(\%)$\downarrow$  \\
			\midrule\midrule
			KF & \textbf{0.5612} & \textbf{0.1681} & \textbf{0.08} & \textbf{0.5776} & \textbf{0.3125} & \textbf{0.42} \\
			
			CV & 0.5515 & 0.1691 & 0.14 & 0.5661 & 0.3159 & 0.58 \\
			KF PD & 0.5516 & 0.1691 & 0.14 & 0.5654 & 0.3158 & 0.63 \\ 
			\bottomrule
		\end{tabular}
	}
	\vspace{-2mm}
	\caption{Comparison of motion models on Waymo Open Dataset. ``KF'' denotes Kalman filters; ``CV'' denotes constant velocity model; ``KF-PD'' denotes the variant using Kalman filter only for motion prediction. Details in Sec.~\ref{subsec::motion_model}.}
	\label{tab::wod_ablate_motion_model}
	\vspace{-4mm}
\end{table}

\begin{table}
	\centering
	\resizebox{0.75\linewidth}{!} 
	{
		\begin{tabular}{@{\hspace{2.0mm}}c|@{\hspace{2.0mm}}c@{\hspace{2.0mm}}c@{\hspace{2.0mm}}c@{\hspace{2.0mm}}c}
			\toprule
			Method & AMOTA$\uparrow$ & AMOTP$\downarrow$ & MOTA$\uparrow$ & IDS $\downarrow$ \\
			\midrule\midrule
			KF & 0.687 & 0.573 & \textbf{0.592} & 519 \\
			CV & \textbf{0.690} & \textbf{0.564} & \textbf{0.592} & \textbf{516} \\
			\bottomrule
		\end{tabular}
	}
	\vspace{-2mm}
	\caption{Comparison of motion models on nuScenes. Abbreviations are identical to Tab.~\ref{tab::wod_ablate_motion_model}. Details in Sec.~\ref{subsec::motion_model}.}
	\label{tab::nuscenes_motion_model}
	\vspace{-4mm}
\end{table}

Motion models depict the motion status of tracklets. They are mainly used to predict the candidate states of objects in the next frame, which are the proposals for the following association step. Furthermore, the motion models like the Kalman filter can also potentially refine the states of objects. 
In general, there are two commonly adopted motion models for 3D MOT: Kalman filter (KF), \eg\ AB3DMOT~\cite{ab3dmot}, and constant velocity model (CV) with predicted speeds from detectors, \eg\ CenterPoint~\cite{center_point}. The advantage of KF is that it could utilize the information from multiple frames and provide smoother results when facing low-quality detection. Meanwhile, CV deals better with abrupt and unpredictable motions with its explicit speed predictions, but its effectiveness on motion smoothing is limited. In Tab.~\ref{tab::wod_ablate_motion_model} and Tab.~\ref{tab::nuscenes_motion_model}, we compare the two of them on WOD and nuScenes, which provides clear evidence for our claims. 

In general, these two motion models demonstrate similar performance. On nuScenes, CV marginally outperforms KF, while it is the opposite on WOD. The advantages of KF on WOD mainly come from the refinement for the bounding boxes. To verify this, we implement the ``KF-PD'' variant, which uses KF only for providing motion predictions prior to association, and the outputs are all original detections. Eventually, the marginal gap between ``CV'' and ``KF-PD'' in Tab.~\ref{tab::wod_ablate_motion_model} supports our claim. On nuScenes, the CV motion model is slightly better due to the lower frame rates on nuScenes (2Hz). To prove our conjecture, we apply KF and CV both under a higher frequency 10Hz setting on nuScenes\footnote{Please check Sec.~\ref{subsec::det_frequency} for how we build 10Hz settings on nuScenes.}, and KF marginally outperforms CV by 0.696 versus 0.693 in AMOTA this time.

To summarize, \emph{the Kalman Filter fits better for high-frequency cases because of more predictable motions, and the constant velocity model is more robust for low-frequency scenarios with explicit speed prediction.} Since inferring the speed is not yet common for detectors, we adopt the Kalman filter for \name\, without loss of generality.

\subsection{Association} 
\label{subsec::association}
\subsubsection{Association Metrics: 3D GIoU}
\label{subsubsec::asso_metrics}
\begin{figure}
	\centering
	\includegraphics[width=1.0\columnwidth]{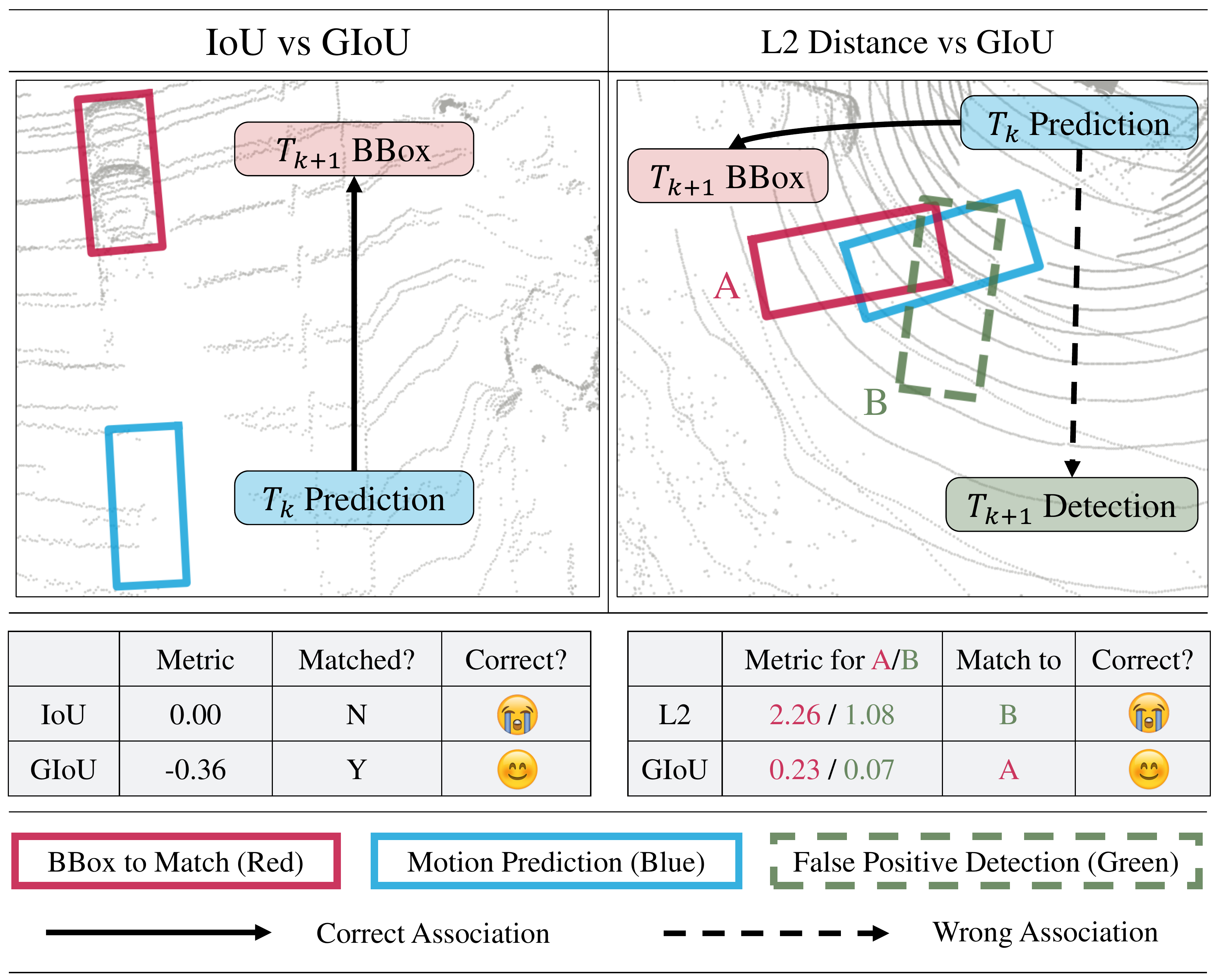}
	\vspace{-2mm}
	\caption{Illustration of association metrics. Left: IoU versus GIoU. Right: L2 Distance versus GIoU. Details are in Sec.~\ref{subsubsec::asso_metrics}.}
	\label{fig::asso_failure}
	\vspace{-4mm}
\end{figure}

\begin{figure*}
    \centering
    \includegraphics[width=1.8\columnwidth]{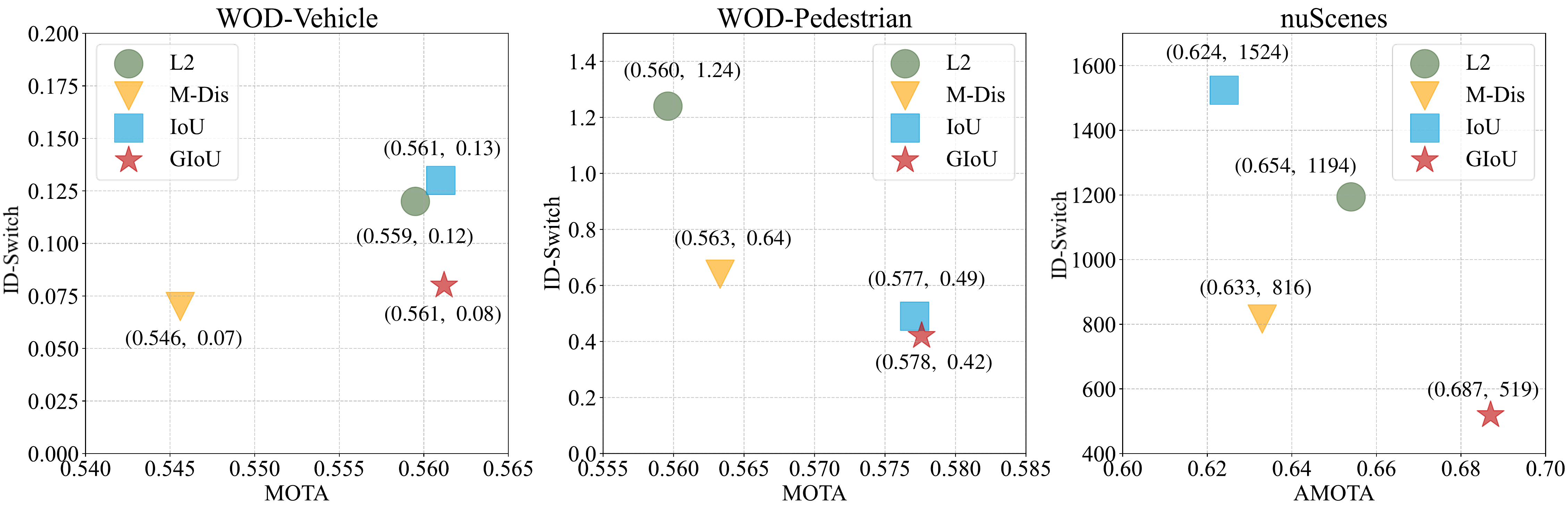}
    \caption{Comparison of association metrics on WOD (left \& middle) and nuScenes (right). ``M-Dis'' is the short for Mahalanobis distance. The best method is closest to the bottom-right corner, having the lowest ID-Switches and highest MOTA/AMOTA.}
    \vspace{-2mm}
    \label{fig::metrics}
    \vspace{-4mm}
\end{figure*}

IoU based~\cite{ab3dmot} and distance based~\cite{mdis, center_point} association metrics are the two prevalent choices in 3D MOT. As in Fig.~\ref{fig::asso_failure}, they have typical but different failure modes. IoU computes the overlapping ratios between bounding boxes, so it cannot connect the detections and motion predictions if the IoU between them are all zeros, which are common at the beginnings of tracklets or on objects with abrupt motions (the left of Fig.~\ref{fig::asso_failure}). 
The representatives for distance-based metrics are Mahalanobis~\cite{mdis} and L2~\cite{center_point} distances. With larger distance thresholds, they can handle the failure cases of IoU based metrics, but they may not be sensitive enough for nearby detection with low quality. We explain such scenarios on the right of Fig.~\ref{fig::asso_failure}. On frame $k$, the blue motion prediction has smaller L2 distances to the green false positive detection, thus it is wrongly associated. Illuminating by such example, we conclude that the distance-based metrics lack discrimination on orientations, which is just the advantage of IOU based metrics.

To get the best of two worlds, we propose to generalize ``Generalized IoU'' (GIoU)~\cite{giou} to 3D for association.
Briefly speaking, for any pair of 3D bounding boxes $B_1, B_2$, their 3D GIoU is as Eq.~\ref{eq::giou}, where $I$, $U$ are the intersection and union of $B_1$ and $B_2$. $C$ is the enclosing convex hull of $U$. $V$ represents the volume of a polygon. We set GIoU $> -0.5$ as the threshold for every category of objects on both WOD and nuScenes for this pair of associations to enter the subsequent matching step.

\begin{equation}
    \label{eq::giou}
        \begin{aligned}
 V_{U} & = V_{B_1} + V_{B_2} - V_{I}, \\
 \mathbf{GIoU}(B_1, B_2) & = V_I / V_U - (V_C - V_U) / V_C.
        \end{aligned}
\end{equation}

As in Fig~\ref{fig::asso_failure}, the GIoU metric can handle both patterns of failures. The quantitative results in Fig.~\ref{fig::metrics} also show the ability of GIoU for improving the association on both WOD and nuScenes.

\subsubsection{Matching Strategies}

Generally speaking, there are two approaches for the matching between detections and tracklets: 1) Formulating the problem as a bipartite matching problem, and then solving it using Hungarian algorithm~\cite{ab3dmot}. 2) Iteratively associating the nearest pairs by greedy algorithm~\cite{mdis, center_point}. 

\begin{figure}
	\centering
	\includegraphics[width=1.0\columnwidth]{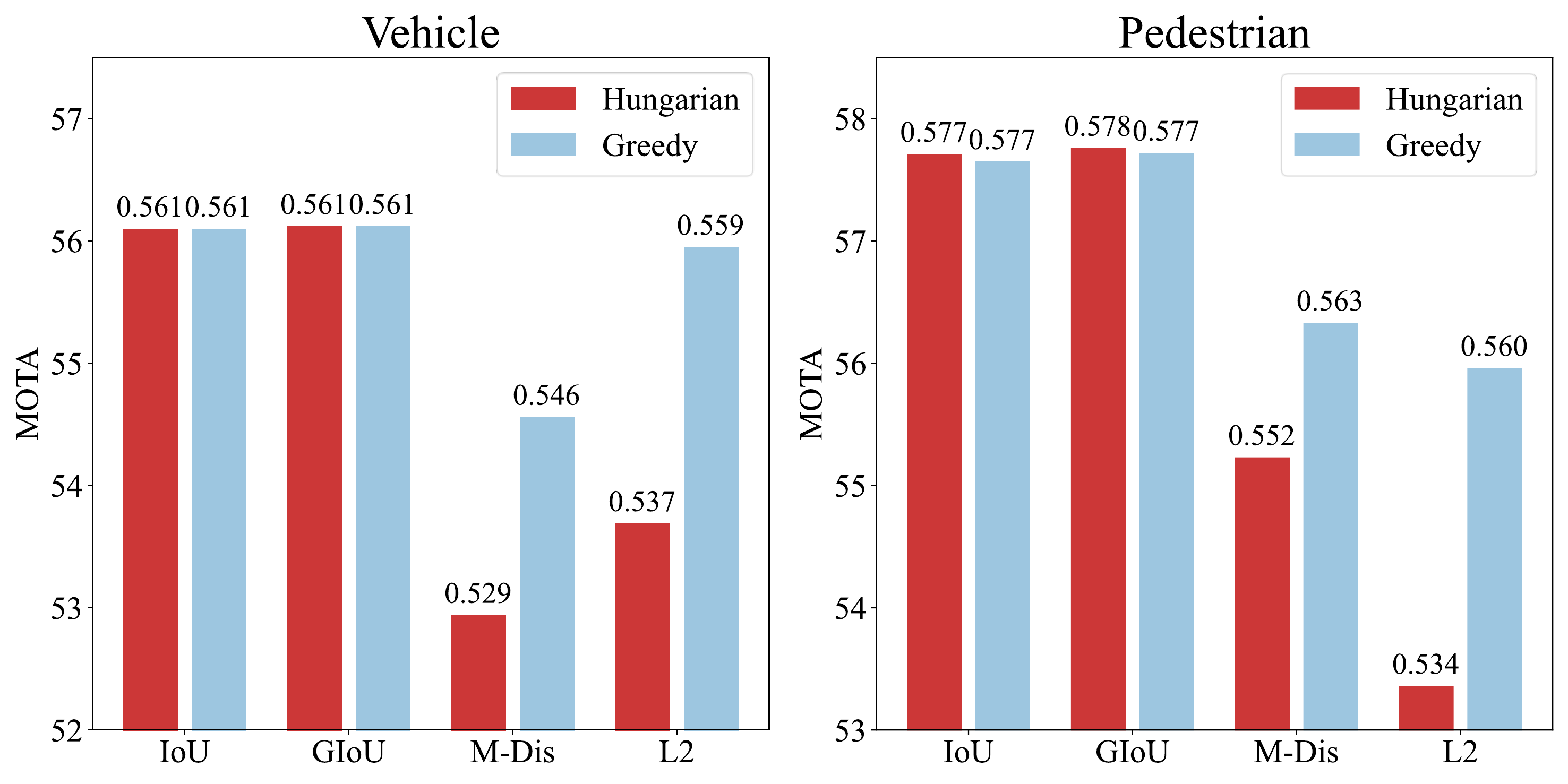}
	\vspace{-3mm}
	\caption{Comparison of matching strategies on WOD.}
	\label{fig::matching_strategy}
	\vspace{-4mm}
\end{figure}

We find that these two methods heavily couples with the association metrics: IoU based metrics are fine with both, while distance-based metrics prefer greedy algorithms. We hypothesize that the reason is that the range of distance-based metrics are large, thus methods optimizing global optimal solution, like the Hungarian algorithm, may be adversely affected by outliers. In Fig.~\ref{fig::matching_strategy}, we experiment with all the combinations between matching strategies and association metrics on WOD. As demonstrated, IoU and GIoU function well for both strategies, while Mahalanobis and L2 distance demand greedy algorithm, which is also consistent with the conclusions from previous work~\cite{mdis}. 

\subsection{Life Cycle Management} 
\label{subsec::life_cycle}

\begin{figure}
    \centering
    \includegraphics[width=1.0\columnwidth]{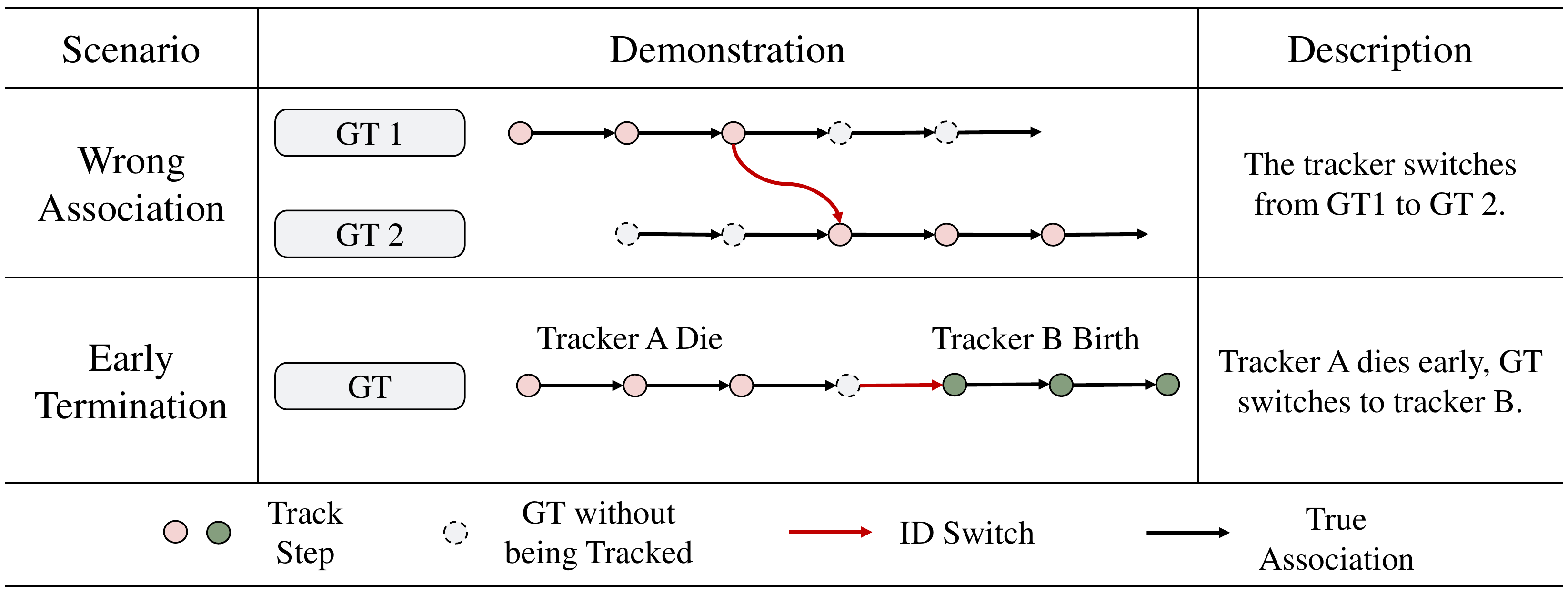}
    \caption{Illustration for two major types of ID-Switches.}
    \label{fig::types_id_switch}
    \vspace{-4mm}
\end{figure}

We analyze all the ID-Switches on WOD\footnote{We use py-motmetrics~\cite{py-motmetrics} for the analysis.}, and categorize them into two groups as in Fig.~\ref{fig::types_id_switch}: wrong association and early termination. Different from the major focus of many work, which is association, we find that the early termination is actually the dominating cause of ID-Switches: 95\% for vehicle and 91\% for pedestrian. Among the early terminations, many of them are caused by point cloud sparsity and spatial occlusion. To alleviate this issue, we utilize the free yet effective information: consensus between motion models and detections with low scores. \emph{These bounding boxes are usually of low localization quality, however they are strong indication of the existence of objects if they agree with the motion predictions.} Then we use these to extend the lives of tracklets.

\begin{figure}
	\centering
	\includegraphics[width=1.0\columnwidth]{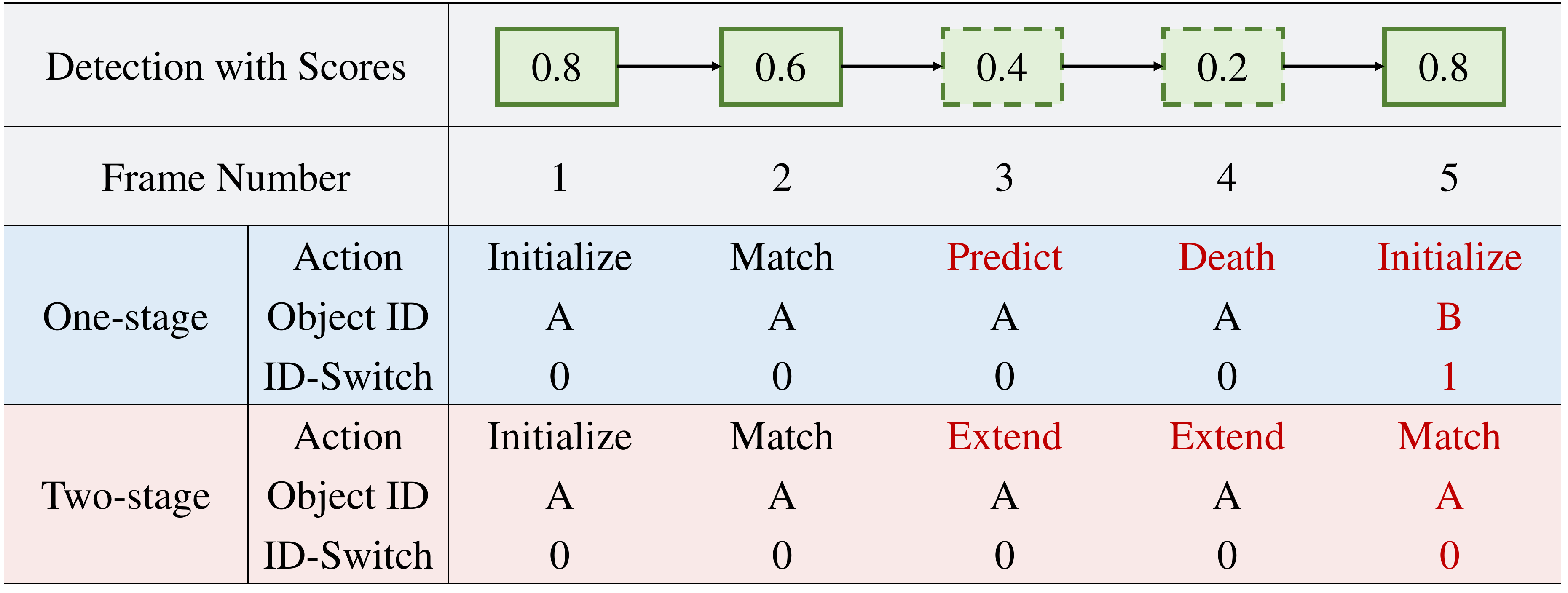}
	\vspace{-4mm}
	\caption{Comparison for ``One-stage'' and ``Two-stage'' association with a hypothetical example. ``Extend'' means ``extending the life cycles,'' and ``Predict'' means ``using motion predictions due to no association.'' Suppose $\mathbf{T}_h=0.5$ and $\mathbf{T}_l=0.1$ are the score thresholds, the ``one-stage'' method early terminates the tracklet because of consecutively lacking associations. Details in Sec.~\ref{subsec::life_cycle}.}
	\label{fig::two_stage}
	\vspace{-2mm}
\end{figure}

Bearing such motivation, we propose ``Two-stage Association.'' Specifically, we apply two rounds of association with different score thresholds: a low one $\mathbf{T}_l$ and a high one $\mathbf{T}_h$ (\eg\ 0.1 and 0.5 for pedestrian on WOD). In stage one, we use the identical procedure as most current algorithms~\cite{ab3dmot, mdis, center_point}: only the bounding boxes with scores higher than $\mathbf{T}_h$ are used for association. In stage two, we focus on the tracklets unmatched to detections in stage one and relax the conditions on their matches: detections having scores larger than $\mathbf{T}_l$ will be sufficient for a match. If the tracklet is successfully associated with one bounding box in stage two, it will still keep being alive. However, as the low score detections are generally in poor quality, we don't output them to avoid false positives, and they are also not used for updating motion models. Instead, we use motion predictions as the latest tracklet states, replacing the low quality detections. 

We intuitively explain the differences between our ``Two-stage Association'' and traditional ``One-stage Association'' in Fig.~\ref{fig::two_stage}. Suppose $\mathbf{T}=0.5$ is the original score threshold for filtering detection bounding boxes, the trackers will then neglect the boxes with scores $0.4$ and $0.2$ on frames $3$ and $4$, which will die because of lacking matches in continuous frames and this eventually causes the final ID-Switch. In comparison, our two-stage association can maintain the active state of the tracklet.

In Tab.~\ref{tab::wod_ablate_life_cycle}, our approach greatly decreases the ID-Switches without hurting the MOTA. This proves that \name\ is effective in extending the life cycles by using detections more flexibly. Parallel to our work, a similar approach is also proven to be useful for 2D MOT~\cite{bytetrack}. 

\begin{table}
    \centering
    \resizebox{1.0\linewidth}{!} 
    {
        \begin{tabular}{@{\hspace{2.0mm}}c|@{\hspace{2.0mm}}c@{\hspace{2.0mm}}c@{\hspace{2.0mm}}c|@{\hspace{2.0mm}}c@{\hspace{2.0mm}}c@{\hspace{2.0mm}}c}
            \toprule
            \multirow{2}{*}{Method}  & \multicolumn{3}{c}{Vehicle} & \multicolumn{3}{c}{Pedestrian} \\
            \cmidrule{2-7} & MOTA$\uparrow$ & MOTP$\downarrow$ & IDS(\%)$\downarrow$  & MOTA$\uparrow$ & MOTP$\downarrow$ & IDS(\%)$\downarrow$  \\
            \midrule\midrule
            One & 0.5567 & 0.1682 & 0.46 & 0.5718 &  \textbf{0.3125} & 0.96  \\
            Two & \textbf{0.5612} & \textbf{0.1681} & \textbf{0.08} & \textbf{0.5776} & \textbf{0.3125} & \textbf{0.42} \\
            \bottomrule
        \end{tabular}
    }
    \caption{Ablation for ``Two-stage Association'' on WOD. ``One'' and ``Two'' denotes the previous one-stage association and our two-stage association methods. Details in Sec.~\ref{subsec::life_cycle}.}
    \label{tab::wod_ablate_life_cycle}
    \vspace{-4mm}
\end{table}

\subsection{Integration of \name}
\label{subsec::summary_improvements}

\begin{figure*}
	\centering
	\includegraphics[width=1.9\columnwidth]{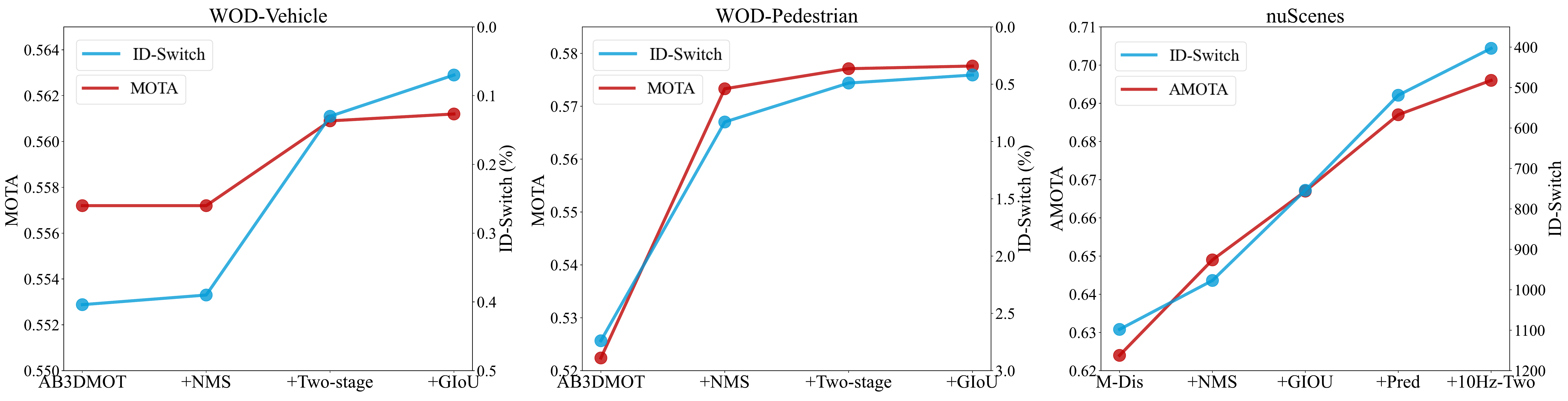}
	\caption{Improvements from \name\ on WOD (left \& middle) and nuScenes (right). We use the common baselines of AB3DMOT~\cite{ab3dmot} on WOD and Chiu \etal~\cite{mdis} on nuScenes. For nuScenes, the improvements of ``10Hz-Two'' (using 10Hz detection and two-stage association) is in Sec.~\ref{subsec::det_frequency}, and ``Pred'' (outputting motion model predictions) is in Sec.~\ref{subsec::rethink_tracklet}. The names for modifications are on the x-axis. Better MOTA and ID-Switch values are higher on the y-axis for clearer visualization.}
	\label{fig::progress}
	\vspace{-4mm}
\end{figure*}

In this section, we integrate the aforementioned techniques into the unified \name\ and demonstrate how they improve the performance step by step. 

In Fig.~\ref{fig::progress}, we illustrate how the performance of 3D MOT trackers improve from the baselines. On WOD, although the properties of vehicles and pedestrian are much different, each technique is applicable to both. On nuScenes, every proposed improvement is also effective for both the AMOTA and ID-Switch.

We also report the test set performance and compare with other 3D MOT methods. Combining our techniques leads to new state-of-the-art results (in Tab.~\ref{tab::wod_test_all}, Tab.~\ref{tab::nusc_test_all}).\footnote{Validation split comparisons are in the supplementary.} 

\begin{table}
    \centering
    \resizebox{1.0\linewidth}{!} 
    {
        \begin{tabular}{@{\hspace{2.0mm}}l|@{\hspace{2.0mm}}c@{\hspace{2.0mm}}c@{\hspace{2.0mm}}c|@{\hspace{2.0mm}}c@{\hspace{2.0mm}}c@{\hspace{2.0mm}}c}
            \toprule
            \multirow{2}{*}{Method} & \multicolumn{3}{c}{Vehicle} & \multicolumn{3}{c}{Pedestrian} \\
            \cmidrule{2-7} & MOTA$\uparrow$ & MOTP$\downarrow$ & IDS(\%)$\downarrow$  & MOTA$\uparrow$ & MOTP$\downarrow$ & IDS(\%)$\downarrow$  \\
            \midrule\midrule
            AB3DMOT~\cite{ab3dmot} & 0.5773 & \best{0.1614} & \second{0.26} & 0.5380 & 0.3163 & \second{0.73} \\
            Chiu \etal~\cite{mdis} & 0.4932 & 0.1689 & 0.62 & 0.4438 & 0.3227 & 1.83 \\
            CenterPoint~\cite{center_point} & \second{0.5938} & 0.1637 & 0.32 & \second{0.5664} & \second{0.3116} & 1.07 \\
            \midrule
            \name & \best{0.6030} & \second{0.1623} & \best{0.08} & \best{0.6013} & \best{0.3114} & \best{0.40} \\
            \bottomrule
        \end{tabular}
    }
\vspace{-2mm}
     \caption{Comparison on WOD test split (L2). CenterPoint~\cite{center_point} detections are used. We mark the best in \best{red} and the second in \second{blue}. We list the methods using public detection. For AB3DMOT~\cite{ab3dmot} and Chiu \etal~\cite{mdis}, we report their best leaderboard entries. }
    \label{tab::wod_test_all}
    \vspace{-2mm}
\end{table}

\begin{table}
    \centering
    \resizebox{0.85\linewidth}{!} 
    {
        \begin{tabular}{@{\hspace{2.0mm}}l|@{\hspace{2.0mm}}c@{\hspace{2.0mm}}c@{\hspace{2.0mm}}c@{\hspace{2.0mm}}c}
            \toprule
            Methods & AMOTA$\uparrow$ & AMOTP$\downarrow$ & MOTA$\uparrow$ & IDS $\downarrow$ \\
            \midrule\midrule
            AB3DMOT~\cite{ab3dmot} & 0.151 & 1.501 & 0.154 & 9027 \\
            Chiu \etal~\cite{mdis} & 0.550 & 0.798 & 0.459 & 776 \\
            CenterPoint~\cite{center_point} & 0.638 & 0.555 & 0.537 & 760 \\
            CBMOT~\cite{score_refinement_3dmot} & 0.649 & 0.592 & 0.545 & \second{557} \\
            OGR3MOT~\cite{graphmot} & 0.656 & 0.620 & 0.554 & \best{288} \\
            \midrule
            \name\ (2Hz) & \second{0.658} & \second{0.568} & \second{0.557} & 609  \\
            \name\ (10Hz) & \best{0.668} & \best{0.550} & \best{0.566} & 575 \\
            \bottomrule
        \end{tabular}
    }
\vspace{-2mm}
    \caption{Comparison on nuScenes test split. CenterPoint~\cite{center_point} detections are used. We list the methods using public detection. We mark the best in \best{red} and the second in \second{blue}. For CBMOT~\cite{score_refinement_3dmot} and OGR3MOT~\cite{graphmot}, we report their numbers with CenterPoint~\cite{center_point} detection. Our numbers using both 2Hz and 10Hz frame rate detections are reported (details of our 10Hz setting are in Sec.~\ref{sec::rethink}). }
    \label{tab::nusc_test_all}
    \vspace{-4mm}
\end{table}


\section{Rethinking nuScenes}
\label{sec::rethink}
Besides the techniques mentioned above, we delve into the design of benchmarks. The benchmarks greatly facilitate the development of research and guide the designs of algorithms. Contrasting WOD and nuScenes, we have the following findings: 1) The frame rate of nuScenes is 2Hz, while WOD is 10Hz. Such low frequency adds unnecessary difficulties to 3D MOT (Sec.~\ref{subsec::det_frequency}). 2) The evaluation of nuScenes requires high recalls with low score thresholds. And it also pre-processes the tracklets with interpolation, which encourages trackers to output the confidence scores reflecting the entire tracklet quality, but not the frame quality (Sec.~\ref{subsec::rethink_tracklet}). We hope these two findings could inspire the community to rethink the benchmarks and evaluation protocols of 3D tracking.

\subsection{Detection Frequencies}
\label{subsec::det_frequency}

Tracking generally benefits from higher frame rates, because motion is more predictable in short intervals. 
We compare the frequencies of point clouds, annotations, and common MOT frame rates on the two benchmarks in Tab.~\ref{tab::freq_comparison}. On nuScenes, it has 20Hz point clouds but only 2Hz annotations. This leads to most common detectors and 3D MOT algorithms work under 2Hz, even they actually utilize all the 20Hz LiDAR data and operate faster than 2Hz. Therefore, we investigate the effect of high-frequency data as follows.

Although the information is more abundant with high frequency (HF) frames, it is non-trivial to incorporate them because nuScenes only evaluates on the low-frequency frames, which we refer to as ``evaluation frames.'' In Tab.~\ref{tab::nuscenes_ablate_detection_freq}, simply using all the 10Hz frames does not improve the performance. This is because the low-quality detection on the HF frames may deviate the trackers and hurt the performance on the sampled evaluation frames. To overcome this issue, we explore by first applying the ``One-stage Association'' on HF frames, where only the bounding boxes with scores larger than $\mathbf{T}_h=0.5$ are considered and used for motion model updating. We then adopt the ``Two-stage Association'' (described in Sec.\ref{subsec::life_cycle}) by using the boxes with scores larger than $\mathbf{T}_l=0.1$ to extend the tracklets. As in Tab.~\ref{tab::nuscenes_ablate_detection_freq}, our approach significantly improves both the AMOTA and ID-Switches. We also try to even increase the frame rate to 20Hz, but this barely leads to further improvements due to the deviation issue. So \name\ uses the 10Hz setting in our final submission to the test set.\footnote{Because of the submission time limits to nuScenes test set, we are only able to report the ``10Hz-One'' variant in Tab.~\ref{tab::nusc_test_all}. It will be updated to ``10Hz-Two'' once we had the chance.}

\begin{table}
	\centering
	\resizebox{0.8\linewidth}{!} 
	{
		\begin{tabular}{@{\hspace{2.0mm}}c|@{\hspace{2.0mm}}c@{\hspace{2.0mm}}c@{\hspace{2.0mm}}c}
			\toprule
			Benchmark & Data & Annotation & Model \\
			\midrule\midrule
			Waymo Open Dataset & 10Hz & 10Hz & 10Hz \\
			nuScenes & 20Hz & 2Hz & 2Hz \\
			\bottomrule
		\end{tabular}
	}
    \vspace{-2mm}
	\caption{Frequency comparison of benchmarks.}
	\label{tab::freq_comparison}
	\vspace{-4mm}
\end{table}

\begin{table}
	\centering
	\resizebox{0.8\linewidth}{!} 
	{
		\begin{tabular}{@{\hspace{2.0mm}}l|@{\hspace{2.0mm}}c@{\hspace{2.0mm}}c@{\hspace{2.0mm}}c@{\hspace{2.0mm}}c}
			\toprule
			Setting & AMOTA$\uparrow$ & AMOTP$\downarrow$ & MOTA$\uparrow$ & IDS $\downarrow$ \\
			\midrule\midrule
			2Hz & 0.687 & 0.573 & 0.592 & 519  \\
			\midrule
			10Hz & 0.687 & 0.548 & 0.599 & 512 \\
			10Hz - One & \textbf{0.696} & 0.564 & \textbf{0.603} & 450 \\
			10Hz - Two & \textbf{0.696} & \textbf{0.547} & 0.602 & \textbf{403} \\
			\midrule
			20Hz - Two & 0.690 & \textbf{0.547} & 0.598 & 416 \\
			\bottomrule
		\end{tabular}
	}
    \vspace{-2mm}
	\caption{MOT with higher frame rates on nuScenes. ``10Hz'' is the vanilla baseline of using all the detections on high frequency (HF) frames. ``-One'' denotes ``One-stage,'' and ``-Two'' denotes ``Two-stage.'' Details in Sec.~\ref{subsec::det_frequency}.}
	\label{tab::nuscenes_ablate_detection_freq}
    \vspace{-4mm}
\end{table}

\subsection{Tracklet Interpolation}
\label{subsec::rethink_tracklet}

The AMOTA metric used in nuScenes calculates the average MOTAR~\cite{ab3dmot} at different recall thresholds, which requires the trackers output the boxes of all score segments. In order to further improve the recall, we output the motion model predictions for frames and tracklets without associated detection bounding boxes, and empirically assign them lower scores than any other detection. In our case, their scores are\ $0.01\times S_P$, where $S_P$ is the confidence score of the tracklet in the previous fram. As shown in Tab.~\ref{tab::nuscenes_ablate_life_cycle}, this simple trick improves the overall recall and AMOTA.

\begin{table}
    \centering
    \resizebox{0.9\linewidth}{!} 
    {
        \begin{tabular}{@{\hspace{2.0mm}}c|@{\hspace{2.0mm}}c@{\hspace{2.0mm}}c@{\hspace{2.0mm}}c@{\hspace{2.0mm}}c@{\hspace{2.0mm}}c}
            \toprule
            Predictions & AMOTA$\uparrow$ & AMOTP$\downarrow$ & MOTA$\uparrow$ & IDS $\downarrow$ & RECALL$\uparrow$\\
            \midrule\midrule
            $\times$ & 0.667 & 0.612 & 0.572 & 754 & 0.696  \\
            \checkmark & \textbf{0.687} & \textbf{0.573} & \textbf{0.592} & \textbf{519} & \textbf{0.725} \\
            \bottomrule
        \end{tabular}
    }
\caption{Improvement from ``outputting motion model predictions'' on nuScenes (2Hz setting).}
\label{tab::nuscenes_ablate_life_cycle}
   \vspace{-4mm}
\end{table}

\begin{figure}
    \centering
    \includegraphics[width=1.0\columnwidth]{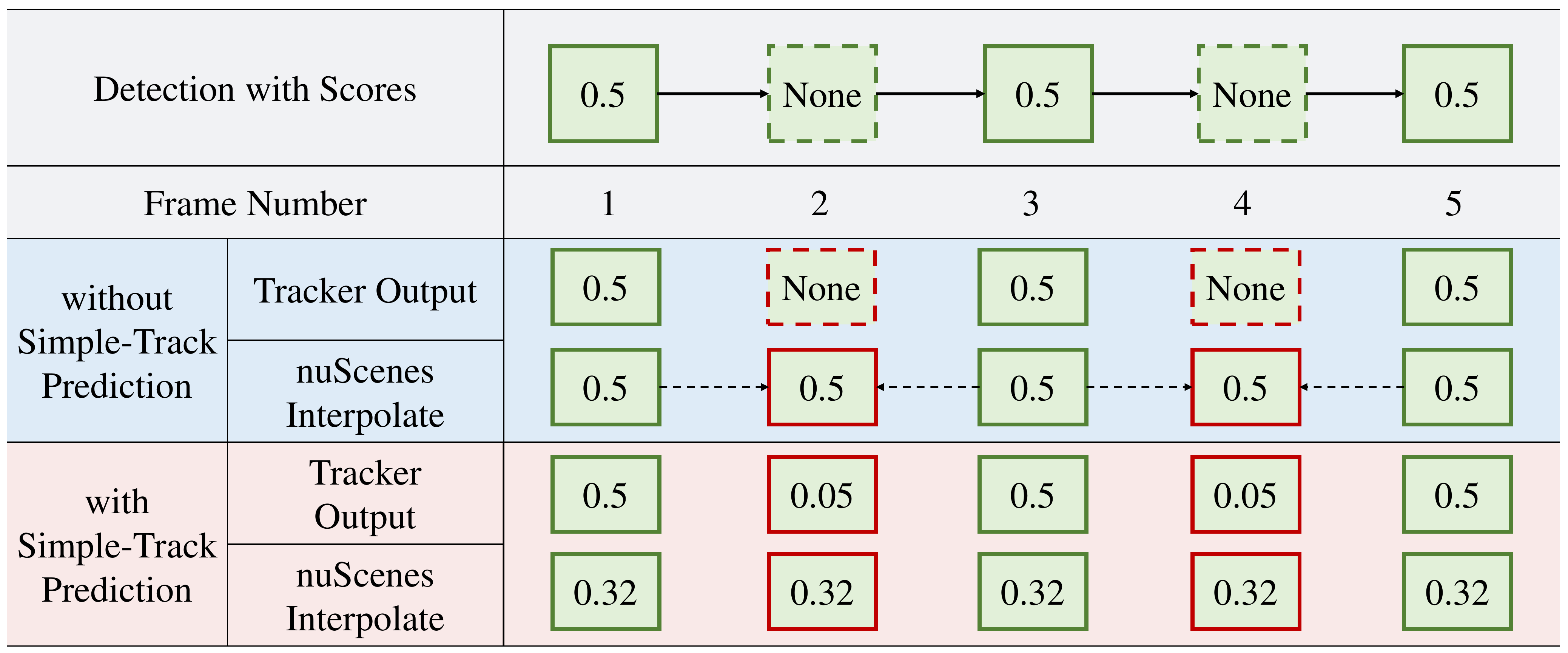}
    \caption{How the motion predictions and nuScenes interpolation changes tracklet scores. Dashed arrows are the directions for interpolation. On Frame 2 and 4 the boxes with score 0.05 are our motion predictions. The ``0.5'' and ``0.32'' are the tracklet-average scores with or without motion predictions. Details in Sec.~\ref{subsec::rethink_tracklet}.}
    \label{fig:score_decrease}
\end{figure}

However, we discover that enhancing the recall is not the only reason for such improvement. Besides the bounding boxes, the scores of the motion model predictions also make a significant contribution. This starts with the evaluation protocol on nuScenes, where they interpolate the input tracklets to fill in the missing frames and change all the scores with their tracklet-average scores as illustrated in Fig.~\ref{fig:score_decrease}. Under this context, our approach can explicitly penalize the low-quality tracklets, which generally contain more missing boxes replaced by motion model predictions. 



In summary, such interpolation on nuScenes encourages the trackers to treat tracklet quality holistically and output calibrated quality-aware scores. However the quality of boxes may vary a lot across frames even for the same tracklet, thus we suggest depicting the quality of a tracklet by only one score is imperfect. Moreover, future information is also introduced in this interpolation step and it changes the tracklet results. This could also bring the concern on whether the evaluation setting is still a fully online one.

\section{Error Analyses}
\label{sec::discussion}

In this section, we conduct analyses on the remaining failure cases of \name\, and propose potential future directions for improving ``tracking by detection'' paradigm. Without loss of generality, we use WOD as an example. 


\subsection{Upper Bound Experiment Settings}
\label{subsec::discussion_upper}

To quantitatively evaluate the causes of failure cases, we contrast \name\ with two different oracle variants.
The results are summarized in Tab.~\ref{tab::wod_upper_bound}.


\noindent
\textbf{GT Output} erases the errors caused by ``output'' policy. We compute the IoU between the bounding boxes from \name\ with the GT boxes at the ``output'' stage, then use the IoU to decide if a box should be output instead of the detection score.  \footnote{The ID-Switch increases because we output more bounding boxes and IDs. The 0.003 false positives in pedestrians are caused by some boxes matching with the same GT box in crowded scenes.}

\noindent
\textbf{GT All} is the upper bound of tracking performance with CenterPoint boxes. We greedily match the detections from CenterPoint to GT boxes, keep the true positive and assign them ground-truth ID.
\begin{table}
    \centering
    \resizebox{1.0\linewidth}{!} 
    {
        \begin{tabular}{@{\hspace{2.0mm}}l|@{\hspace{2.0mm}}c@{\hspace{2.0mm}}c@{\hspace{2.0mm}}c@{\hspace{2.0mm}}c|@{\hspace{2.0mm}}c@{\hspace{2.0mm}}c@{\hspace{2.0mm}}c@{\hspace{2.0mm}}c}
            \toprule
            \multirow{2}{*}{Method} & \multicolumn{4}{c}{Vehicle} & \multicolumn{4}{c}{Pedestrian} \\
            \cmidrule{2-9} & MOTA$\uparrow$ & IDS(\%)$\downarrow$ & FP$\downarrow$ & FN$\downarrow$  & MOTA$\uparrow$ & IDS(\%)$\downarrow$ & FP$\downarrow$ & FN$\downarrow$ \\
            \midrule\midrule
            \name\ & 0.561 & 0.078 & 0.104 & 0.334 & 0.578 & 0.425 & 0.109 & 0.309 \\
            GT Output & 0.741 & 0.104 & \textbf{0.000} & 0.258 & 0.778 & 0.504 & 0.003 & 0.214 \\
            GT All & \textbf{0.785} & \textbf{0.000} & \textbf{0.000} & \textbf{0.215} & \textbf{0.829} & \textbf{0.000} & \textbf{0.000} & \textbf{0.171} \\ 
            \bottomrule
        \end{tabular}
    }
    \vspace{-3mm}
    \caption{Oracle Experiments on WOD. }
    \label{tab::wod_upper_bound}
    \vspace{-4mm}
\end{table}

\subsection{Analyses for ``Tracking By Detection''}
\noindent
\textbf{ID-Switches.} We break down the causes of ID-Switches as in Fig.~\ref{fig::types_id_switch}. 
Although early termination has been greatly decreased by the scale of 86\% for vehicle and 70\% for pedestrian with ``Two-stage Association,'' it still takes up 88\% and 72\% failure cases in the remaining ID-Switches in \name\ for vehicle and pedestrian, respectively. We inspect these cases and discover that most of them result from long-term occlusion or the returning of objects from being temporarily out of sight. Therefore, in addition to improving the association, potential future work can develop appearance models like in 2D MOT~\cite{wojke2017simple, leal2016learning, sadeghian2017tracking, li2020graph} or silently maintain their states to re-identify these objects after they are back.

\noindent
\textbf{FP and FN.} The ``GT All'' in Tab.~\ref{tab::wod_upper_bound} shows the upper bound for MOT with CenterPoint~\cite{center_point} detection, and we analyze the class of vehicle for example. Even with ``GT All'' the false negatives are still 0.215, which are the detection FN and can hardly be fixed under the ``tracking by detection'' framework. Comparing ``GT All'' and \name, we find that the tracking algorithm itself introduces 0.119 false negatives. We further break them down as follows. Specifically, the difference between ``GT Output" and ``GT ALL" indicates that the 0.043 false negatives are caused by the uninitialized tracklets resulting from NMS and score threshold in pre-processing. The others come from life-cycle management. The ``Initialization'' requires two frames of accumulation before outputting a tracklet, which is same as AB3DMOT~\cite{ab3dmot}. This yields a marginal 0.005 false negatives. Our ``Output'' logic uses detection score to decide output or not, taking up the false negatives number 0.076. 
Based on these analyses, we can conclude that the gap is mainly caused by the inconsistency between the scores and detection quality.
By using historical information, 3D MOT can potentially provide better scores compared to single frame detectors, and this has already drawn some recent attention~\cite{graphmot, score_refinement_3dmot}. 





\section{Conclusions and Future Work}

In this paper, we decouple the ``tracking by detection'' 3D MOT algorithms into several components and analyze their typical failures. With such insights, we propose corresponding enhancements of using \textit{NMS}, \textit{GIoU}, and \textit{Two-stage Association}, which lead to our \name. In addition, we also rethink the frame rates and interpolation pre-processing in nuScenes. We eventually point out several possible future directions for ``tracking by detection'' 3D MOT.

However, beyond the ``tracking by detection'' paradigm, there are also branches of great potential. For better bounding box qualities, 3D MOT can refine them using long term information~\cite{qi2021offboard, auto4d, pang_lidarsot}, which are proven to outperform the detections based only on local frames. The future work can also transfer the current manual rule-based methods into learning-based counterparts, \eg\ using learning based intra-frame mechanisms to replace the NMS, using inter-frame reasoning to replace the 3D GIoU and life cycle management, etc.


\paragraph{Acknowledgment.} We would like to thank Tianwei Yin for kindly helping us during our applying the CenterPoint detection to 3D multi-object tracking.

\renewcommand\thesection{\Alph{section}}
\renewcommand\thetable{\Alph{table}}  
\setcounter{section}{0}
\setcounter{table}{0}
\section{Appendix for SimpleTrack}
\subsection{Validation Split Comparison}
\label{sec::validation}
We compare our \name\ with other 3D MOT methods on the validation splits as in Tab.~\ref{tab::wod_validation_all} and Tab.~\ref{tab::nuscenes_validation_all}. In the experiments, our \name\ also demonstrates strong performance. On both Tab.~\ref{tab::wod_validation_all} and Tab.~\ref{tab::nuscenes_validation_all}, our \name\ can outperform the methods without learning based modules, which is consistent with the test set performance in the main paper (Tab.~\ref{tab::wod_test_all} and Tab.~\ref{tab::nusc_test_all}). In addition, we find it interesting in Tab.~\ref{tab::nuscenes_validation_all} that the learning based method OGR3MOT~\cite{graphmot} can achieve better performance than our 2Hz \name, which demonstrate the potential of applying learning techniques for 3D MOT. However, such advantage of OGB3MOT vanishes for AMOTA on the test set, as in the Tab.~\ref{tab::nusc_test_all} of the main paper. This suggests that our learning-free modifications may have the ability to adapt to the domain gaps in the data.

%
%
%

\subsection{Experimental Setup}

Due to the space constraints, we discuss the detailed hyper-parameters and settings for our \name\ here.

\paragraph{Waymo Open Dataset}
\vspace{-3mm}
\begin{enumerate}[itemsep=-1mm]
	\item \textbf{Pre-process.} We use CenterPoint detection~\cite{center_point}, and then apply NMS with the IoU threshold equals to 1/4 onto the detection bounding boxes.
	
	\item \textbf{Association.} We use GIoU as the association metric and Hungarian algorithm to solve the matchings. The threshold for GIoU is -0.5 across all types of objects. 
	
	\item \textbf{Motion Model.} We use the default Kalman filter parameters as AB3DMOT~\cite{ab3dmot}, and pair the usages of Hungarian algorithm.
	
	\item \textbf{Life Cycle Management.} The life cycle management is the same as AB3DMOT~\cite{ab3dmot}, 3 hits to start outputting a tracker and consecutive 2 misses terminates a tracklet. We set the threshold for outputting detection bounding boxes as 0.7 for vehicle and cyclist, and 0.5 for pedestrian. In our ``Two-stage association,'' we adopt the low score threshold as 0.1. 
\end{enumerate}

\begin{table}
	\centering
	\resizebox{1.0\linewidth}{!} 
	{
		\begin{tabular}{@{\hspace{2.0mm}}l|@{\hspace{2.0mm}}c@{\hspace{2.0mm}}c@{\hspace{2.0mm}}c|@{\hspace{2.0mm}}c@{\hspace{2.0mm}}c@{\hspace{2.0mm}}c}
			\toprule
			\multirow{2}{*}{Method} & \multicolumn{3}{c}{Vehicle} & \multicolumn{3}{c}{Pedestrian} \\
			\cmidrule{2-7} & MOTA$\uparrow$ & MOTP$\downarrow$ & IDS(\%)$\downarrow$  & MOTA$\uparrow$ & MOTP$\downarrow$ & IDS(\%)$\downarrow$  \\
			\midrule\midrule
			AB3DMOT$^{*}$~\cite{ab3dmot} & \second{0.5572} & \best{0.1679} & 0.40  & 0.5224 & \second{0.3098} & 2.74 \\
			Chiu \etal$^{*}$~\cite{mdis} & 0.5406 & 0.1665 & 0.37 & 0.4810 & \best{0.3086} & 3.34 \\
			CenterPoint~\cite{center_point} & 0.5505 & 0.1691 & \second{0.26} & \second{0.5493} & 0.3137 & \second{1.13} \\
			\midrule
			\name\ & \best{0.5612} & \second{0.1681} & \best{0.08} & \best{0.5776} & 0.3125 & \best{0.42} \\
			\bottomrule
		\end{tabular}
	}
	\caption{Results on WOD validation split (L2). We mark the best in \best{red} and the second in \second{blue}. For fair comparison, we list the methods using the public CenterPoint~\cite{center_point} detection, $^*$ means the numbers from our own implementations.}
	\label{tab::wod_validation_all}
\end{table}

\begin{table}
	\centering
	\resizebox{0.85\linewidth}{!} 
	{
		\begin{tabular}{@{\hspace{2.0mm}}l|@{\hspace{2.0mm}}c@{\hspace{2.0mm}}c@{\hspace{2.0mm}}c@{\hspace{2.0mm}}c}
			\toprule
			Methods & AMOTA$\uparrow$ & AMOTP$\downarrow$ & MOTA$\uparrow$ & IDS $\downarrow$ \\
			\midrule\midrule
			AB3DMOT$^{*}$~\cite{ab3dmot} & 0.598 & 0.771 & 0.537 & 1570 \\
			AB3DMOT~\cite{ab3dmot}\cite{graphmot} & 0.578 & 0.807 & 0.514 & 1275 \\
			Chiu \etal$^{*}$~\cite{mdis} & 0.624 & 0.655 & 0.542 & 1098 \\
			Chiu \etal~\cite{mdis}\cite{graphmot} & 0.617 & 0.984 & 0.533 & 680 \\
			CenterPoint~\cite{center_point} & 0.665 & \second{0.567} & 0.562 & 562 \\
			CBMOT~\cite{score_refinement_3dmot} & 0.675 & 0.591 & 0.583 & 494 \\
			\midrule
			MPN-Baseline~\cite{graphmot} & 0.593 & 0.832 & 0.514 & 1079 \\
			OGR3MOT~\cite{graphmot} & \second{0.693} & 0.627 & 0.602 & \best{262} \\
			\midrule
			\name\ (2Hz) & 0.687 & 0.573 & \best{0.592} & 519 \\
			\name\ (10Hz) & \best{0.696} & \best{0.547} & \second{0.602} & \second{405} \\
			\bottomrule
		\end{tabular}
	}
	\caption{Results on nuScenes validation set. We mark the best in \best{red} and the second in \second{blue}. For fair comparison, we list the methods using public CenterPoint~\cite{center_point} detection, the numbers marked with $^*$ are our own implementations, the numbers marked with~\cite{graphmot} are from OGR3MOT~\cite{graphmot}.}
	\vspace{-4mm}
	\label{tab::nuscenes_validation_all}
\end{table}

\paragraph{nuScenes}
\vspace{-3mm}
\begin{enumerate}[itemsep=-1mm]
    \item \textbf{Pre-process.} We apply NMS according to IoU threshold  equal to 1/10. After NMS, all the remaining detections are kept as the input to 3D MOT algorithms.
    \item \textbf{Association.} We adopt the same settings as on Waymo Open Dataset.
    \item \textbf{Motion Model.} The settings for our motion model is identical to that on Waymo Open Dataset.
    \item \textbf{Life Cycle Management.} We adopt the similar strategy as the tracking algorithm in Center Point~\cite{center_point}, where the trackers start outputting upon the first association, and are terminated after two continuous misses.
\end{enumerate}

{\small
	\bibliographystyle{ieee_fullname}
	\bibliography{reference}
}
\end{document}